\documentclass[10pt,twocolumn,letterpaper]{article}

\usepackage[pagenumbers]{cvpr} 
\usepackage{booktabs}
\usepackage{multirow}
\usepackage{diagbox}
\usepackage{pgfplots}
\usepackage{tcolorbox}
\usepackage{verbatim}
\usepackage{amsmath}
\usepackage{amssymb}
\newcommand{\mypara}[1]{\vspace{0.5em}\noindent\textbf{#1}}
\newcommand{\ignore}[1]{}

\definecolor{cvprblue}{rgb}{0.21,0.49,0.74}
\usepackage[pagebackref,breaklinks,colorlinks,allcolors=cvprblue]{hyperref}

\def\paperID{*****} 
\def\confName{CVPR}
\def\confYear{2025}

\title{AVC-DPO: Aligned Video Captioning via Direct Preference Optimization}

\author{
    \textbf{Jiyang Tang$^{2}$\thanks{Equal contribution.}}~,
  \textbf{Hengyi Li$^{3*}$}, \textbf{Yifan Du$^{1*}$} \textbf{Wayne Xin Zhao$^{1}$\thanks{Correspondence to Wayne Xin Zhao.}}~
  \\
  $^1$Gaoling School of Artificial Intelligence, Renmin University of China\\
  $^2$College of Artificial Intelligence, Nankai University\\
  $^3$School of Computer Science and Technology, Beijing Institute of Technology\\
  \texttt{\{tjyaiccc,liharry18,yifandu1999,batmanfly\}@gmail.com}
}

\begin{document}
\maketitle
\begin{abstract}

Although video multimodal large language models (video MLLMs) have achieved substantial progress in video captioning tasks, it remains challenging to adjust the focal emphasis of video captions according to human preferences. To address this limitation, we propose Aligned Video Captioning via Direct Preference Optimization (AVC-DPO), a post-training framework designed to enhance captioning capabilities in video MLLMs through preference alignment. Our approach designs enhanced prompts that specifically target temporal dynamics and spatial information—two key factors that humans care about when watching a video—thereby incorporating human-centric preferences. AVC-DPO leverages the same foundation model's caption generation responses under varied prompt conditions to conduct preference-aware training and caption alignment. Using this framework, we have achieved exceptional performance in the LOVE@CVPR'25 Workshop Track 1A: Video Detailed Captioning Challenge, achieving first place on the Video Detailed Captioning (VDC) benchmark according to the VDCSCORE evaluation metric.
\end{abstract}    
\section{Introduction}
\label{sec:intro}


Recent advances demonstrate that video MLLMs exhibit outstanding performance across diverse video understanding tasks~\cite{zhang2024video,chen2025versavid,peng2025actionart}, progressively advancing toward general video intelligence. Video captioning is an important task that requires the model to generate comprehensive and coherent textual descriptions given a video~\cite{chai2024auroracap,chen2024personalized,zhang2025vcapsbench}. The challenge of this task mainly lies in capturing complex visual information within individual frames and modeling temporal dependencies across multiple frames.

Current video MLLMs have improved video captioning primarily through two approaches:
One is optimizing the reasoning mechanism, which enhances the logicality and completeness of generated captions \cite{ghazanfari2025chain,meng2025videocapr1enhancingmllmsvideo}. The other is spatio-temporal fine-grained control, integrating visual representations with temporal information to enable detailed descriptions of dynamic scenes \cite{tang2025caption,lian2025describe,xue2025progress}. However, these methods predominantly focus on improving the model's comprehensive understanding of video content, overlooking the practical utility of captions for human preference. Specifically, they lack alignment with human preferences on content emphasis in descriptive text.

To address this issue, we introduce Aligned Video Captioning via Direct Preference Optimization (AVC-DPO), a new framework to align video MLLMs to generate rich, structured captions across diverse aspects. Our methodology first employs prompt enhancement to augment the model's capability in generating aspect-specific captions, followed by a principle-based scoring mechanism to evaluate caption quality against predefined criteria. Finally, we construct preference pairs from scored outputs to optimize caption generation through Direct Preference Optimization~(DPO). We evaluate AVC-DPO on the VDC benchmark \cite{chai2024auroracap}, demonstrating effective improvements in caption quality across five key aspects: camera, short, background, main object, and detailed. 

The major contributions of our work are as follows:
\begin{itemize}
\item We propose a pipeline that automatically synthesizes video caption preference pairs from various aspects, without requiring human annotation.

\item Based on this framework, we significantly improve the captioning capability of Qwen2.5-VL-7B, achieving first place in the LOVE@CVPR’25 Workshop Track 1A: Video Detailed Captioning Challenge.
\end{itemize}

\section{Method}
\label{sec:Method}
\begin{figure*}
    \centering
    \includegraphics[width=1\linewidth]{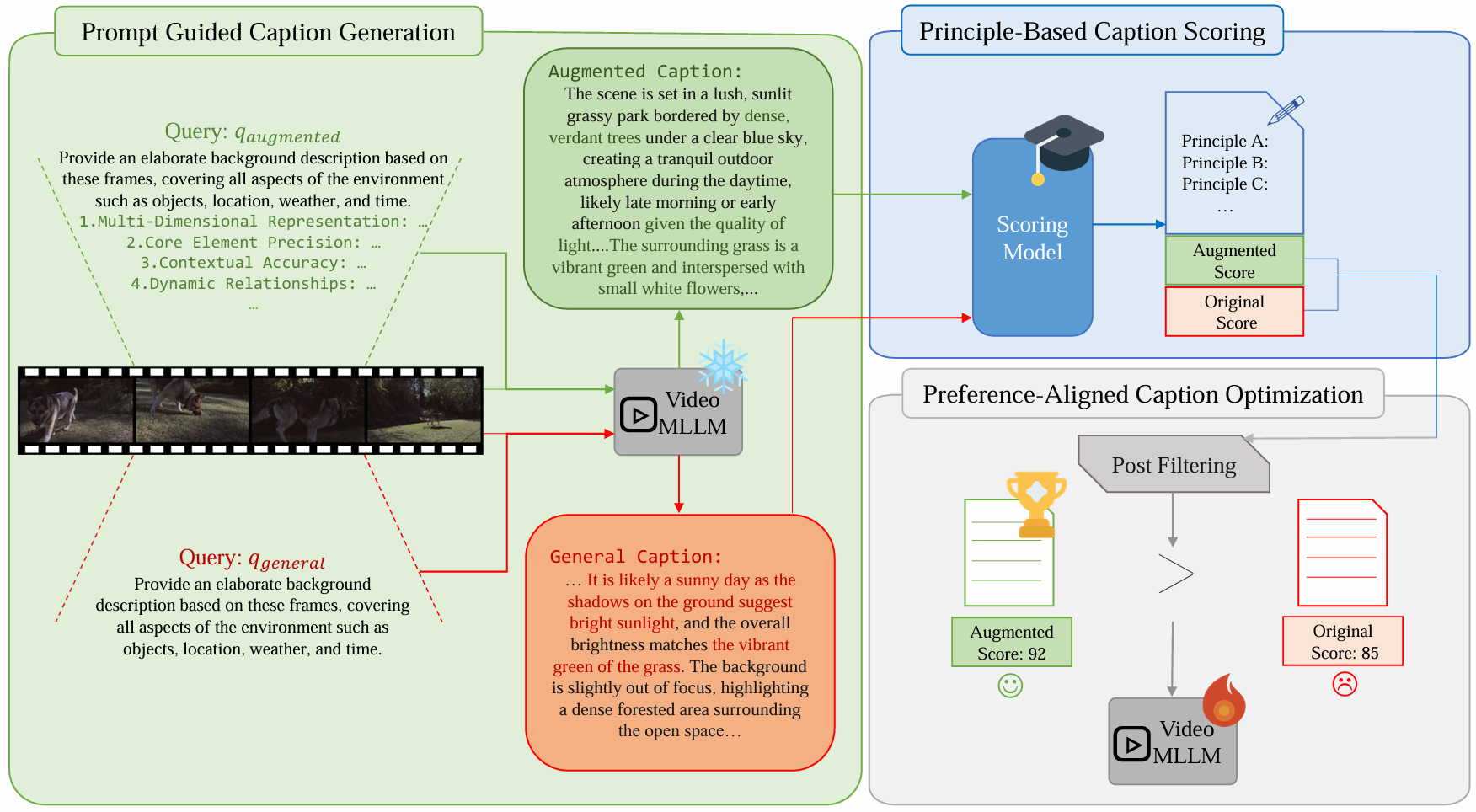}
    \caption{Aligned Video Captioning via Direct Preference Optimization (AVC-DPO) automatically synthesizes video caption preference pairs through three core stages:
(1)Prompt-Guided Caption Generation(left): Aspect-specific prompts enhance baseline model outputs;
(2)Principle-Based Caption Scoring(upper-right): Scoring models design scoring principles to evaluate captions under general vs. augumented prompts;
(3)Preference-Aligned Caption Optimization(lower-right): Cases with significant quality improvements in augumented captions are selected to optimize output alignment.}
    \label{fig:pipeline}
\end{figure*}
In this section, we introduce AVC-DPO, a framework for automatically synthesizing video caption preference pairs from multiple perspectives. The framework consists of three key stages: (1) \textit{Prompt-Guided Caption Generation}, which designs prompts to steer the model toward generating captions aligned with specific aspects; (2) \textit{Principle-Based Caption Scoring}, which leverages a stronger model to define scoring principles and evaluate the generated captions; and (3) \textit{Preference-Aligned Caption Optimization}, which selects positive and negative samples for subsequent alignment training. The overall architecture of AVC-DPO is depicted in Figure~\ref{fig:pipeline}.

\subsection{Preliminaries}
\label{subsec:Preliminaries}
video MLLM usually consists of a large language model (LLM) $f_\theta$, a visual encoder $f_\phi$, and a multimodal projector $f_\psi$. These models process video inputs $V$ and textual query $Q$. The input video is first tokenized into visual tokens by the visual encoder, then projected into the text embedding space via the multimodal projector $\bold{x}_v = f_\psi(f_\phi(V))$, while the query is encoded into textual embedding $\bold{x}_t$. This framework models the probability of response $\bold{y}$:
\begin{equation}
P(\bold{y}|\bold{x}_t,\bold{x}_v) = \prod_{i=1}^n f_\theta (y_i|y_{<i}, \bold{x}_t,\bold{x}_v).
\end{equation}
Preference learning aims to align model outputs with human preferences, one of its most prevalent methods is DPO. To apply DPO for preference training in video MLLMs, for each video input $V$ and query $Q$ in the preference data, there exists a preferred output $\bold{y}^+$ and a dispreferred output $\bold{y}^-$ satisfying $P(\bold{y}^+|\bold{x}_t,\bold{x}_v) > P(\bold{y}^-|\bold{x}_t,\bold{x}_v)$.

\subsection{Prompt-Guided Caption Generation}
\label{subsec:Prompt-Guided Caption Generation}
To synthesize caption preference pairs, we focus on two fundamental dimensions of video understanding: temporal and spatial. Temporal understanding captures dynamic changes and event progression over time, while spatial understanding focuses on the static visual layout and object relationships within a frame. These complementary aspects are essential for fine-grained comprehension of video content. To highlight each dimension, we construct aspect-specific prompts, denoted as $\bold{p}^{\text{temp}}$ and $\bold{p}^{\text{spa}}$, which guide the model to generate captions that emphasize temporal or spatial features, respectively. Each is combined with a base query $\bold{q}_0$ to form an aspect-augmented query: $\bold{q}^{\text{temp}} = \bold{q}_0||\bold{p}^{\text{temp}}, \bold{q}^{\text{spa}} = \bold{q}_0||\bold{p}^{\text{spa}}$. We then query the video MLLM with $\bold{q}^{\text{temp}}$ and $\bold{q}^{\text{spa}}$ to obtain positive samples, which are expected to reflect fine-grained, aspect-specific understanding. In contrast, negative samples are generated using only the general query $\bold{q}_0$, leading to more generic video descriptions. This contrast ensures that the preference pairs clearly distinguish between general and aspect-specific comprehension.

\subsection{Principle-Based Caption Scoring}
\label{subsec:Principle-Based Caption Scoring}
Following the prompt-guided caption generation phase, we obtain candidate preference pairs for each video across different aspects. However, some of these pairs may be unreliable, for example, cases where the dispreferred caption is comparable to or better than the preferred one, or where the preferred caption is misaligned with the intended query. To mitigate such noise and improve data quality, we adopt a principle-based caption scoring method inspired by \citet{liu2025inference}. Specifically, we employ a stronger MLLM (\ie Qwen2.5-VL-72B) to assess caption quality. This model is prompted with evaluation principles, and is used to score captions generated under both the base query $\bold{q}_0$ and the aspect-augmented queries $\bold{q}^{\text{temp}}/\bold{q}^{\text{spa}}$. To filter high-quality preference pairs, we retain only those whose score difference exceeds a predefined threshold $\delta = 5$. Each final pair is denoted as $(V,\bold{q},\bold{y}^+,\bold{y}^-)$, where $V$ is the video, $\bold{q}=\bold{q}_0$ is the general query, $\bold{y}^+$ is the caption generated under the augmented query (\eg $\bold{q}^{\text{temp}}$ or $\bold{q}^{\text{spa}}$), and $\bold{y}^-$ is the caption generated under the base query $\bold{q}_0$. This filtering step ensures that retained pairs reflect a clear and meaningful preference aligned with the targeted aspect.


\subsection{Preference-Aligned Caption Optimization}
\label{subsec:Preference-Aligned Caption Optimization}
Direct Preference Optimization (DPO) \cite{rafailov2023direct} is adopted due to its flexibility and stability in preference-based learning. Given the dataset collected in Section~\ref{subsec:Principle-Based Caption Scoring} $\mathcal{D}_{\text{DPO}} = \{(V, \bold{q}, \bold{y}^+, \bold{y}^-)\}$ and the video MLLM $f_\theta$, the loss function is defined as:
\begin{equation}
\begin{aligned}
L&_{\text{DPO}}(f_{\theta};f_{\mathrm{ref}}) = -\mathbb{E}_{(V,\bold{q},\bold{y}^{+},\bold{y}^{-}) \sim \mathcal{D}} \\
&\Bigg[ \log \sigma \Bigg( \beta \bigg( \log \frac{f_{\theta}(\bold{y}^{+}|V,\bold{q})}{f_{\mathrm{ref}}(\bold{y}^{+}|V,\bold{q})}
 - \log \frac{f_{\theta}(\bold{y}^{-}|V,\bold{q})}{f_{\mathrm{ref}}(\bold{y}^{-}|V,\bold{q})} \bigg) \Bigg) \Bigg]
\end{aligned}
\end{equation}
where $\sigma$ is the sigmoid function. This objective drives the model to assign higher probabilities to preferred outputs, aligning its behavior more closely with human preference.

\section{Experiments}
\begin{table*}[t]
\centering
\small
\begin{tabular}{l|cccccc}
\toprule
\multicolumn{1}{c|}{Model}                                                        & \begin{tabular}[c]{@{}c@{}} Camera\\ (Acc / Score)\end{tabular} & \begin{tabular}[c]{@{}c@{}}  Short \\ (Acc / Score)\end{tabular} & \begin{tabular}[c]{@{}c@{}}  Background\\ (Acc / Score)\end{tabular} & \begin{tabular}[c]{@{}c@{}} Main Object\\ (Acc / Score)\end{tabular} & \begin{tabular}[c]{@{}c@{}} Detailed\\ (Acc / Score)\end{tabular} & \begin{tabular}[c]{@{}c@{}} Average\\ (Acc / Score)\end{tabular} \\
\midrule
 Aria-3.5B×8~\citep{li2024aria}
& 42.4/2.2& 33.2/1.8& 40.9/2.1& 48.9/2.5& 47.3/2.4&42.5/2.2\\
 LLaVA-Video-7B~\citep{zhang2024llava-vid}& 46.1/2.3& 32.8/1.7& 37.6/1.9& 46.2/2.4& 35.0/1.8&39.0/2.0\\
 VideoChat-Flash-7B~\citep{li2024videochat}
& 43.7/2.3& 33.7/1.7& 45.1/2.3& 47.6/2.4& 44.5/2.3&42.9/2.2\\
 Cockatiel-8B(Distilled)~\citep{qin2025cockatielensemblingsynthetichuman}
& 42.3/2.2& \textbf{44.0/2.3}& 43.9/2.3& 43.9/2.3& 44.0/2.3&43.6/2.3\\
 Cockatiel-13B~\citep{qin2025cockatielensemblingsynthetichuman}
& 42.6/2.2& \underline{43.5/2.3}& 44.1/2.3& 44.4/2.3& 44.4/2.3&43.8/2.3\\
 VideoCap-R1~\citep{meng2025videocapr1enhancingmllmsvideo}
& -& -& -& -& -&43.8/2.4\\
 Qwen2-VL-7B~\citep{wang2024qwen2-vl}& 39.0/2.1& 36.8/1.9& 41.9/2.1& 47.8/2.5& 42.5/2.2&41.6/2.2\\
 \midrule
 Qwen2.5-VL-7B~\citep{Qwen2.5-VL}& 42.5/2.3& 36.7/1.9& 45.2/2.3& 49.0/2.5& 46.1/2.4&43.9/2.3\\
 AVC-DPO-temporal-7B& \underline{50.4/2.7}& 39.0/2.0& \underline{49.9/2.6}& \underline{50.5/2.6}& \underline{48.9/2.5}&\underline{47.7$\textcolor{blue}{(+3.8)}$/2.5$\textcolor{blue}{(+0.2)}$}\\
 AVC-DPO-spatial-7B& \textbf{52.7/2.8}& 39.8/2.1& \textbf{54.7/2.8}& \textbf{55.1/2.8}& \textbf{53.3/2.8}&\textbf{51.1$\textcolor{blue}{(+7.2)}$/2.6$\textcolor{blue}{(+0.3)}$}\\
\bottomrule
\end{tabular}

\caption{
Quantitative Comparison of existing VDC models on VDCSCORE Benchmark. 
We report the score of baseline methods based on the official \textsc{AuroraCap} website. The best and second-best results are emphasized using \textbf{bold} and \underline{underline}.
}
\label{tab:vdc_benchmark_results}
\end{table*}

\subsection{Experimental Setup}

\mypara{Training Dataset.}
To synthesize preference pairs for DPO, we select ShareGPT4Video~\cite{chen2024sharegpt4videoimprovingvideounderstanding} as the data source due to its broad domain coverage. We filter out the video that appears in the evaluation benchmark and obtain 711 \textit{spatial} preference pairs and 1,280 \textit{temporal} preference pairs.

\mypara{Evaluation Benchmarks.}
We evaluate our model on VDC\cite{chai2024auroracap}, a video detailed captioning benchmark comprising over 1,000 meticulously annotated structured captions. This benchmark enables rigorous assessment of the quality of detailed video descriptions. We follow the official evaluation method and use VDCSCORE as the metric to evaluate model performance.

\mypara{Baseline Methods.}
We compare our method with two categories of models: (1) video captioning MLLMs, including Cockatiel~\cite{qin2025cockatielensemblingsynthetichuman} and VideoCap-R1~\cite{meng2025videocapr1enhancingmllmsvideo}; and (2) general video MLLMs at the top of the VDC benchmark leaderboard, including VideoChat-Flash-7B~\cite{li2024videochat}, LLaVA-Video-7B~\cite{zhang2024llava-vid}, and Aria-3.5B~\cite{li2024aria}.
\ignore{
\begin{itemize}
    \item \textbf{Cockatiel\cite{qin2025cockatielensemblingsynthetichuman}}: Cockatiel-13B leverages a learned scorer to curate high-quality captions for fine-grained, human-preferred video descriptions and is then trained to inherit ensemble strengths while aligning with human preferences.
    \item \textbf{VideoCap-R1\cite{meng2025videocapr1enhancingmllmsvideo}}:
    VideoCap-R1 uses a GRPO-based reinforcement learning framework with dual rewards—an LLM-free reasoning scorer and an LLM-assisted caption scorer—to link structured video analysis and caption generation, significantly improving action description accuracy.
    \item \textbf{Aria-3.5B\cite{li2024aria}}:
    Aria-3.5B is a native multimodal MoE model achieving state-of-the-art results in video and document understanding, supporting a 64K-token context window with efficient inference and low-cost fine-tuning by activating only 3.9B parameters per token.
    \item \textbf{VideoChat-Flash-7B\cite{li2024videochat}}:
    VideoChat-Flash-7B employs hierarchical visual token compression (HiCo) and a multi-stage short-to-long training strategy to efficiently model long videos with high-fidelity features, achieving state-of-the-art performance on long- and short-video benchmarks at the 7B scale.
    \item \textbf{LLaVA-Video-7B\cite{zhang2024llava-vid}}:
    LLaVA-Video-7B is trained on LLaVA-Video-178K and LLaVA-OneVision Dataset, based on Qwen2 language model with a context window of 32K tokens.
\end{itemize}
}

\mypara{Implementation Details.}
For DPO training, we adopt the LLaMA-Factory framework~\cite{zheng2024llamafactory}. During training, the ViT is frozen, while both the projector and the LLM are trainable. All experiments are conducted with a batch size of 4, using 1 training epoch, gradient accumulation steps set to 2. \ignore{and a cosine learning rate schedule with linear warmup from 0 to 1e-5 over the first 10\% of training steps.} For video inputs, we sample at 2 fps and limit the maximum number of frames to 32. All experiments are carried out using 8$\times$A800-80GB GPUs.




\subsection{Experimental Results}
Table \ref{tab:vdc_benchmark_results} shows that our models, AVC-DPO-temporal-7B and AVC-DPO-spatial-7B, achieve state-of-the-art performance on the VDC benchmark, outperforming models such as Cockatiel-13B, LLaVA-Video-7B, and Aria-3.5B×8 across four key dimensions: camera, background, main object, and detailed caption. Notably, AVC-DPO-spatial-7B reaches an average score of 54.0\%, surpassing Cockatiel-8B (Distilled) at 43.5\%. These results underscore our models' strengths in spatial and temporal video understanding.

\begin{figure}[b]\centering
\includegraphics[width=0.98\linewidth]{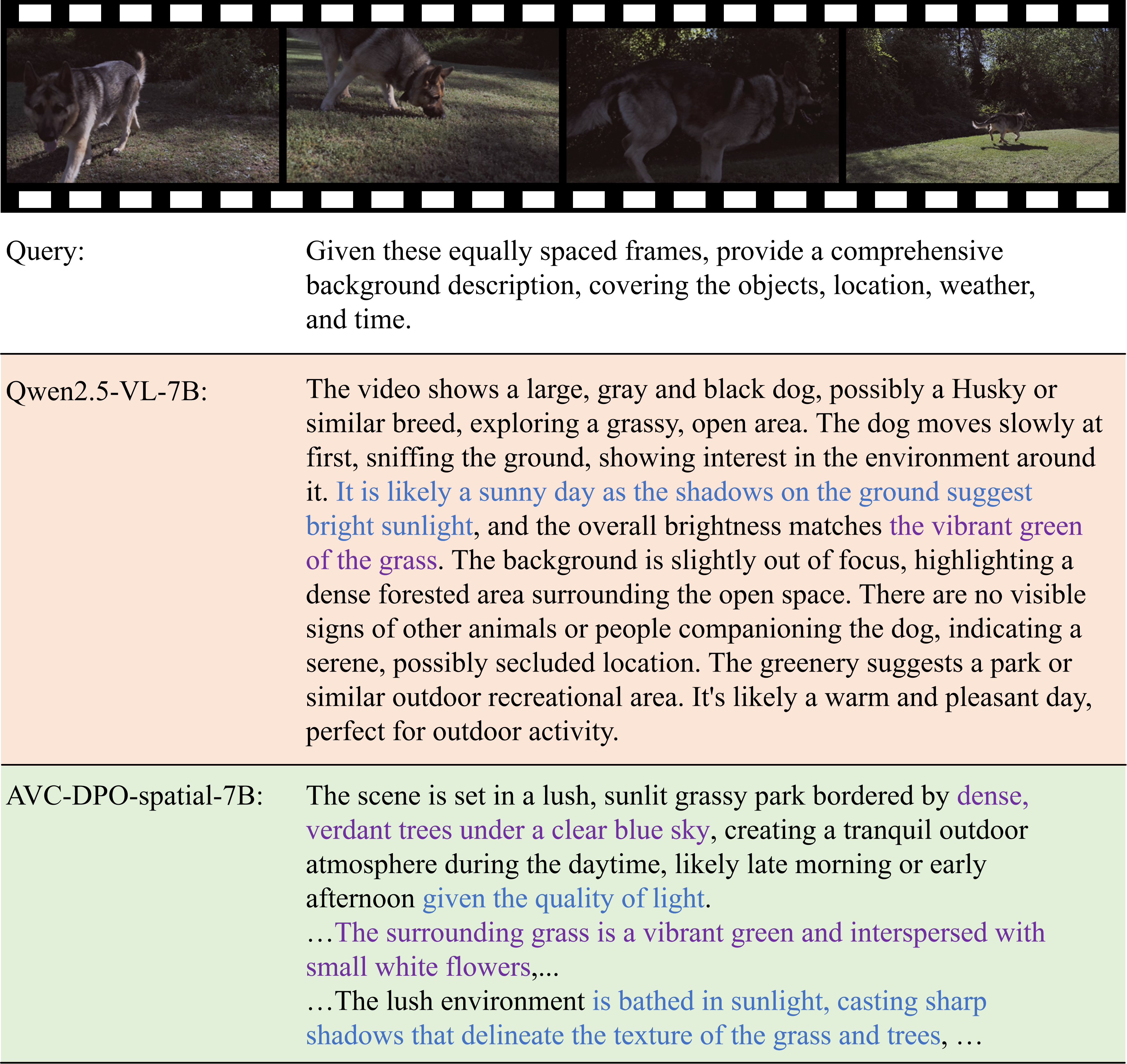}
\caption{
Case study comparing AVC-DPO-spatial-7B with Qwen2.5-VL-7B. Blue text highlights lighting-related descriptions; purple text indicates spatial entity recognition.
}

\label{fig:case study}
\end{figure}

Figure~\ref{fig:case study} presents a qualitative comparison between AVC-DPO-spatial-7B and Qwen2.5-VL-7B. Notable improvements are observed in two dimensions. First, \textit{lighting-aware spatial modeling} is significantly enhanced: AVC-DPO-spatial-7B offers detailed descriptions like ``sunlit grassy park'' and ``late morning or early afternoon'', compared to the base model's vague mentions of ``bright sunlight''. Second, in \textit{spatial entity grounding and relational topology}, the base model provides a general description of the environment, it identifies specific visual elements (\eg ``small white flowers'', ``dense, verdant trees'') and their layout, unlike the base model’s general descriptions. These improvements highlight the effectiveness of our AVC-DPO method in enhancing detailed video captioning.

However, performance gaps remain in short captioning tasks due to an inherent conflict in our DPO training between detail and brevity. Optimizing for rich spatial and temporal cues leads to longer outputs that violate the “one-sentence” constraint. For example, AVC-DPO-spatial-7B produces captions averaging 679.78 tokens. This stems from preference pairs that reward informativeness without considering length. Future work could address this trade-off by incorporating length constraints into DPO objectives.

\subsection{Ablation Study}



\begin{table}
  \centering
  \begin{tabular}{l|cccccc}
\toprule
\multicolumn{1}{c|}{Model}                                & \begin{tabular}[c]{@{}c@{}}  Background Acc \end{tabular} & \begin{tabular}[c]{@{}c@{}} Background Score\end{tabular} \\
\midrule
 Qwen2.5-VL-7B
& 45.2 & 2.3 \\
 + spatial prompt
& 49.2 & 2.5 \\
 + spatial DPO
& 54.7 & 2.8 \\
\bottomrule
\end{tabular}
  \caption{
    Ablation on the spatial dimension using \textsc{Background} scores. The first row is the 7B base model; the second uses spatial-enhanced prompts at inference; the third applies spatial DPO with the original prompt. Results indicate that prompt enhancement alone is insufficient to match DPO performance.
    }
  \label{tab:ablation_results}
\end{table}

We design specific prompts in Section~\ref{sec:Method} to guide the video MLLM in generating detailed captions from various aspects. In this section, we ablate whether prompt engineering alone can improve model performance. The results in \Cref{tab:ablation_results} show that spatial prompts yield only modest improvements (around 4.0\%), whereas spatial DPO training achieves significantly stronger performance (around 9.5\%). This demonstrates that prompt engineering alone is insufficient to match the effectiveness of DPO training.

\section{Conclusion}
In this paper, we propose AVC-DPO, a simple yet effective post-training framework that enhances a MLLM’s video-captioning capabilities. Our method automatically synthesizes video-caption preference pairs for alignment training. On the Video Detailed Caption (VDC) benchmark, AVC-DPO delivers substantial gains over the base model and secures first place in the LOVE@CVPR’25 Workshop Track 1A: Video Detailed Captioning Challenge.

{
    \small
    \bibliographystyle{ieeenat_fullname}
    \bibliography{main}
}


\end{document}